%% file: main.tex
\documentclass[10pt,twocolumn,letterpaper]{article}

\usepackage[pagenumbers]{cvpr} 

\usepackage[title]{appendix}

\usepackage{graphicx}
\usepackage{amsmath}
\usepackage{amssymb}
\usepackage{booktabs}
\usepackage{tabularx}
\usepackage{makecell}
\usepackage[table]{xcolor}
\usepackage{lipsum}  
\usepackage{dsfont}
\usepackage{mathrsfs}
\usepackage{xcolor}
\usepackage[pagebackref,breaklinks,colorlinks]{hyperref}

\usepackage[capitalize]{cleveref}
\crefname{section}{Sec.}{Secs.}
\Crefname{section}{Section}{Sections}
\Crefname{table}{Table}{Tables}
\crefname{table}{Tab.}{Tabs.}
\newcommand{\Acronym}[0]{VEDet\xspace}
\newcommand{\Quat}[0]{\mathbf{\bar{q}}}
\newcommand\mypar[1]{\par\vspace{2.0mm}\noindent\textbf{#1}\;\;}

\def\cvprPaperID{6940} 
\def\confName{CVPR}
\def\confYear{2023}

\begin{document}

\title{Viewpoint Equivariance for Multi-View 3D Object Detection}

\author{Dian Chen \quad Jie Li \quad Vitor Guizilini \quad Rareș Ambruș \quad Adrien Gaidon\\
Toyota Research Institute (TRI), Los Altos, CA\\
{\tt\small \{firstname.lastname\}@tri.global}
}
\maketitle

\begin{abstract}
\input{sections/0_abstract.tex}
\end{abstract}

\section{Introduction}
\label{sec:intro}
\input{sections/1_introduction.tex}

\section{Related Work}
\label{sec:related}
\input{sections/2_related.tex}

\section{Viewpoint Equivariant 3D Detection}
\label{sec:method}
\input{sections/3_method.tex}

\section{Experiments}
\label{sec:experiments}
\input{sections/4_experiments.tex}

\vspace{-2mm}
\section{Limitations}
\input{sections/5_limitations.tex}
\label{sec: limitation}
\vspace{-2mm}
\section{Conclusion}
\label{sec:conclusion}
\input{sections/6_conclusion.tex}

\section*{Appendix}
\begin{appendix}
\section{Implementation details}
\mypar{VEDet model.} We use three different backbones to report performance on NuScenes: ResNet-50 and ResNet-101 \cite{he2016deep} are initialized from the ImageNet-pretrained weights hosted on OpenMMLab \cite{mmcv}; VoVNetV2-99 \cite{lee2020centermask} is initialized from the depth-pretrained weights released by \cite{park2021pseudo}. The image features and geometric positional encodings have dimension $C=256$, and are added element-wise as the keys to the transformer decoder, which has $L=6$ transformer layers. In the transformer layers, we use multi-head attention with $8$ heads, dropout rate $0.1$ on the residual connection, and $2048$ hidden dimensions in the feed-forward network. To predict the classification scores, we use a single linear projection from $256$-dim queries to $10$-dim class scores; for predicting the 3D box attributes, we use a $2$-layer MLP with $[512, 512]$ hidden dimensions  interleaved with ReLU activations. The classification and regression heads are both shared across the $6$ transformer layers.

\mypar{Learnable geometry mapping.} For the MLP in the learnable geometry mapping, used to make both geometric positional encoding and object queries, we use $1$ hidden layer with $1920$ dimensions, followed by a ReLU activation and a final projection to $C=256$ dimensions. Therefore, given Fourier bands $k=64$, the dimensions go through the following changes: $d_0 \rightarrow_{\textrm{Fourier}} 1280 \rightarrow_{\textrm{hidden}} 1920 \rightarrow_{\textrm{proj}} 256$, where $d_0=10$ for both perspective geometry of an image feature $[\mathbf{r}_{(u_i,v_i)}, \Quat,\mathbf{t}]$ and query geometry $[\mathbf{c}_j^v,\Quat^v,\mathbf{t}^v]$.

\mypar{Query points.} We use $900$ learnable 3D query points in all experiments. We follow \cite{wang2022detr3d} to use object ranges $[-51.2m, -51.2m, -5.0m, 51.2m, 51.2m, 3.0m]$ in XYZ axes of the global BEV space around the vehicle. The query points are normalized to $[0, 1]$ by a sigmoid operation and scaled by their range. The predictions of box center offsets are added to the points before the sigmoid operation.

\mypar{Virtual view sampling.} During training, the range we use to uniformly sample the translation for the virtual query views is $[-0.6m, -1.0m, -0.3m, 0.6m, 1.0m, 0m]$ in XYZ axes. We uniformly sample the yaw angle to be between $[0, 2\pi]$.

\mypar{Temporal modeling.} In the full-version \Acronym we concatenate $2$ temporal frames at the token dimension. Following \cite{huang2022bevdet4d,liu2022petrv2}, we randomly sample one frame from the past $[3, 27]$ frames during training, and use the past $15$-th frame during inference. The time interval between consecutive frames is roughly $0.083$s.

\mypar{Optimization.} During training, the loss weights we use are $\lambda_{cls}=2.0$ and $\lambda_{reg}=0.25$ following \cite{wang2022detr3d,liu2022petr}. We use the AdamW optimizer \cite{loshchilov2017decoupled} with weight decay $0.01$. The learning rate is linearly warmed up in the first $500$ iterations from $6.77e^{-5}$ ($\frac{1}{3}$ of initial learning rate) to $2e^{-4}$. The learning rate of the pretrained backbone is multiplied by $0.1$ compared to all other components, that are trained from scratch. Checkpointing \cite{chen2016training} is adopted during training to save GPU memory, bringing the training time of the full-version \Acronym (2 frames, $640\times1600$ images, $V=2$) to 20 hours on 8 A100 GPUs, for 24 epochs on NuScenes.

\mypar{Data augmentation.} We use data augmentations following \cite{liu2022petr}, in the order shown below:

\begin{itemize}
    \item Resize. The original images are resized keeping the aspect ratio. The resize factor is sampled uniformly from $[0.564, 0.8]$ for $384\times1056$ images, $[0.79, 1.1]$ for $512\times1408$ images, and $[0.94, 1.25]$ for $640\times1600$ images.
    \item Crop. Given a crop size $H\times W$ and an intermediate image size $H'\times W'$ after the resizing, the top area $[0, H'-H]$ is cropped to meet the final height $H$. The left limit of the cropping box is uniformly sampled from $[0, W'-W]$.
    \item Horizontal flip. With a $50\%$ probability, we flip all $N$ images at the same time, alongside the 3D box annotations. The camera poses and intrinsics are transformed accordingly to reflect the flipping. Concretely, the X coordinate of the camera translation and yaw angle are flipped, while the principal point in the intrinsic matrix has the X-coordinate flipped.
    \item Global rotation. Without changing the images, the camera poses and 3D box annotations are rotated around the Z axis of the global BEV space. The angle is uniformly sampled from $[-22.5^{\circ}, 22.5^{\circ}]$.
    \item Global scaling. Without changing the images, the camera poses and 3D box annotations are scaled relative to the origin of the global BEV space. The scaling factor is uniformly sampled from $[0.95, 1.05]$.
\end{itemize}

During testing, no random augmentations are used. The images are resized to the final width while keeping the aspect ratio, and cropped at the bottom-center.

\end{appendix}

{\small
\bibliographystyle{ieee_fullname}
\bibliography{egbib}
}

\end{document}

%% file: sections/0_abstract.tex
3D object detection from visual sensors is a cornerstone capability of robotic systems. State-of-the-art methods focus on reasoning and decoding object bounding boxes from multi-view camera input. In this work we gain intuition from the integral role of multi-view consistency in 3D scene understanding and geometric learning. To this end, we introduce \Acronym, a novel 3D object detection framework that exploits 3D multi-view geometry to improve localization through viewpoint awareness and equivariance. VEDet leverages a query-based transformer architecture and encodes the 3D scene by augmenting image features with positional encodings from their 3D perspective geometry. We design view-conditioned queries at the output level, which enables the generation of multiple virtual frames during training to learn viewpoint equivariance by enforcing multi-view consistency. The multi-view geometry injected at the input level as positional encodings and regularized at the loss level provides rich geometric cues for 3D object detection, leading to state-of-the-art performance on the nuScenes benchmark. The code and model are made available at \href{https://github.com/TRI-ML/VEDet.git}{https://github.com/TRI-ML/VEDet}.

%% file: sections/1_introduction.tex
\input{figures/fig1-teaser.tex}
Camera-based 3D object detection is a critical research topic, with important applications in areas such as autonomous driving and robotics due to the semantic-rich input and low cost compared to range sensors.
In the past few years, monocular 3D detection has seen significant progress, from relying on predicting pseudo point clouds as intermediate representation \cite{ wang2019pseudo,weng2019monocular,qian2020end} to end-to-end learning~\cite{simonelli2019disentangling,park2021pseudo,wang2021fcos3d}. 
However, monocular 3D detectors are inherently ambiguous in terms of depth, which motivated some recent exploration in multi-view and multi-sweep 3D object detection~\cite{li2022bevformer,wang2022detr3d,liu2022petr,liu2022petrv2}. 

In a conventional monocular setting, given multiple cameras on a sensor rig, single-view detections are merged to the global frame through rule-based processing such as Non-Maximum Suppression (NMS). Recent advances in multi-view camera-based 3D algorithms~\cite{wang2022detr3d, liu2022petr} proposed to jointly aggregate multi-view information at the feature level, and directly predict a single set of detections in the global frame. These algorithms demonstrate a giant leap in 3D detection performance on multi-camera datasets (E.g., Nuscenes~\cite{caesar2020nuscenes}).
To aggregate information from different views, one line of query-based detectors adopt transformers to query image features \cite{wang2022detr3d,liu2022petr,liu2022petrv2} or bird's-eye-view (BEV) features \cite{li2022bevformer,jiang2022polarformer} via an attention mechanism. In contrast, another line of works ``lift-splat-shoot'' \cite{philion2020lift} image features from each view into the shared BEV features to be processed by convolutional detection heads \cite{li2022bevdepth}.

To further mitigate the depth ambiguity, some concurrent works have started extending multi-view to ``multi-sweep'' across timestamps and observe a promising performance boost~\cite{li2022bevformer,liu2022petrv2}.

While the works mentioned above demonstrate a strong potential for multi-view 3D detection, progress has concentrated on input aggregation and information interplay across frames and less on learning objectives. We argue that the learning objective can play a crucial role in ingesting the core knowledge in a multi-view setting: 3D geometry. 

This paper proposes to encourage 3D geometry learning for multi-view 3D detection models through viewpoint awareness and equivariance. 
We obtain our intuition from traditional structure-from-motion works~\cite{build-rome}, where multi-view geometry is modeled through multi-view consistency.
To this end, we propose viewpoint-awareness on the object queries, as well as a multi-view consistency learning objective as a 3D regularizer that enforces the model to reason about geometry.
Compared to existing methods that make 3D predictions in the default egocentric view, our proposed multi-view predictions and viewpoint equivariance effectively bring stronger geometric signals conducive to the 3D reasoning.
More specifically, in our query-based framework, the geometry information of image features and object queries is injected completely via implicit geometric encodings, and the transformer decoder is expected to learn better correspondence and 3D localization under the viewpoint equivariance objective.
We demonstrate that our proposed framework can make the best of available geometry information with extensive experiments and establish the new state-of-the-art in multi-view 3D object detection.
In summary, our contributions are:
\begin{itemize}
    \item  We propose a novel \textbf{Viewpoint Equivariance (VE)} learning objective that encourages multi-view consistency in 3D detection models, leading to improved 3D object detection performance.
    \item We propose a new multi-view 3D object detection framework, \textbf{\Acronym}, which employs a query-based transformer architecture with perspective geometry and viewpoint awareness injected both at the encoding and decoding stages. \Acronym fully enables our proposed VE learning objective, facilitating geometry learning with implicit inductive biases.
    \item \Acronym achieves \textbf{state-of-the-art on large-scale benchmark}, reaching \textbf{45.1\%mAP on NuScenes val set and 50.5\% mAP on test set}. We provide a comprehensive analysis of our components, and share insights based on empirical observations.
\end{itemize}

%% file: figures/fig1-teaser.tex
\begin{figure}[]
    \centering
    \includegraphics[width=0.97\linewidth]{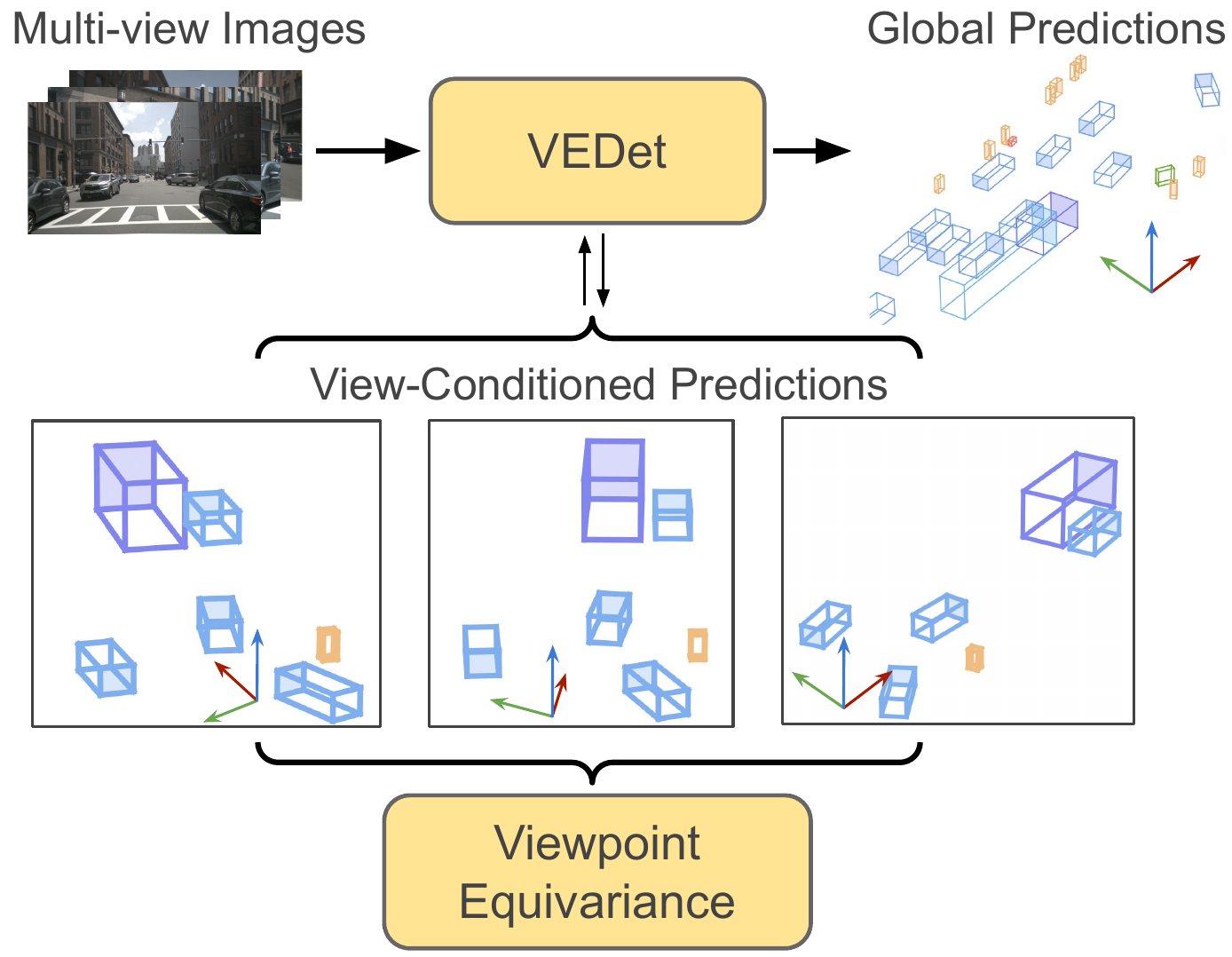}
    \caption{
    \textbf{Our proposed \Acronym} encodes the 3D scene from multi-view images, and decodes objects with view-conditioned queries. The predicted 3D bounding boxes are expressed in the underlying views of the queries, which enables us to enforce viewpoint equivariance among predictions from multiple views. Virtual query views are generated during training and together with the viewpoint equivariance regularization bring richer geometric learning signals to guide the model to better understand the 3D structure in the scene. During inference, the global predictions can be obtained by simply choosing the global frame as the query view. 
    }
    \label{fig:teaser}
\end{figure}

%% file: sections/2_related.tex
\subsection{Monocular 3D object detection}
Early works tackled camera-based 3D object detection in a monocular setting by adopting a two-stage pseudo-LiDAR paradigm \cite{wang2019pseudo,weng2019monocular,qian2020end,you2019pseudo} or directly building upon 2D detection frameworks to predict extra 3D properties \cite{mousavian20173d,xu2018multi,brazil2019m3d,simonelli2019disentangling,park2021pseudo,wang2021fcos3d,chen2022epro}. Due to the inherent scale ambiguity in depth estimation, one standard approach was to lift from 2D to 3D by aligning 3D properties and their 2D projections on the image plane \cite{barabanau2019monocular,he2019mono3d++,li2019gs3d,ku2019monocular}, while others leveraged additional object or scene priors such as shapes \cite{he2019mono3d++}, CAD models \cite{chabot2017deep}, or ground planes \cite{ansari2018earth}. In the line of representation learning, DD3D \cite{park2021pseudo} exploited large-scale pre-training to learn a depth-aware representation that can universally benefit 3D detection algorithms. However, all these methods still struggle with the two major drawbacks in monocular 3D detection: inherent depth ambiguity and insufficient context to infer objects across images. As a multi-view method, our work addresses both of these issues by leveraging the ample 3D geometric cues in multi-camera setups. 

\subsection{Multi-view 3D object detection}
Recent advances in camera-based 3D object detection have started to leverage multi-view context, which can improve the detection of objects that appear in more than one image. One line of works extends the DETR framework \cite{carion2020end}, which decodes 3D bounding boxes with a set of queries \cite{wang2022detr3d,liu2022petr,liu2022petrv2,li2022bevformer,jiang2022polarformer,li2022unifying}. Using camera parameters, DETR3D \cite{wang2022detr3d} directly projects 3D queries to 2D image planes to update query features, while PETR \cite{liu2022petr} constructs 3D position embeddings from point frustums to implicitly guide query updates. BEVFormer \cite{li2022bevformer} and UVTR \cite{li2022unifying} first build an intermediate voxelized feature space around the vehicle/robot's ego coordinate frame, before feeding the features to a DETR-style decoder. Another line of work follows LSS \cite{philion2020lift} and constructs the voxelized feature space, before applying a detection head on the features to predict bounding boxes. Typically, a depth head is also trained to predict a depth bin distribution in order to lift-splat-shoot the features, as in BEVDepth \cite{li2022bevdepth}. Our work falls in the first line of research and exploits multi-view geometric consistency to improve bounding box localization in 3D space.

\subsection{Implicit geometric encoding}
The Transformer architecture \cite{vaswani2017attention,dosovitskiy2020image,han2022survey} introduced the use of positional encodings for input features. This brought upon a paradigm shift in how to model the relative position of elements, from \textit{explicitly}, i.e. by recurrent operations or convolutional filters, to \textit{implicitly}, i.e. learned from data. Inspired by this new paradigm, some works started to investigate how to use positional encodings constructed from geometric priors as input-level inductive biases \cite{yifan2022input,liu2022petr,liu2022petrv2}. ILIB \cite{yifan2022input} utilizes multi-view geometry, including camera and epipolar cues, to generate position encodings, to be processed by a generalist Perceiver IO~\cite{jaegle2021perceiver} architecture to produce a set of latent vectors. From this latent space, ILIB constructs queries from 3D viewing rays to decode depth maps. PETR \cite{liu2022petr,liu2022petrv2} similarly constructs position encodings from generated point frustums at pre-defined depth values, and decodes 3D bounding boxes using queries constructed from 3D anchor points \cite{wang2022anchor} in the ego vehicle space. Positional encoding has also been used extensively in the context of neural fields (i.e. coordinate-based multi-layer perceptron) to process the appearance, radiance, or occupancy of a scene~\cite{sitzmann2019siren,mildenhall2021nerf,xie2021neural}. Our work improves the design of 3D geometric position encoding and introduces a new optimization objective to guide the detection model toward learning better object localization.

%% file: sections/3_method.tex
The multi-view 3D object detection task aims at detecting 3D bounding boxes in the scene with class labels, given a set of images from $N$ cameras with poses and intrinsics $\{\mathbf{I}_i \in \mathbb{R}^{3\times H\times W},\mathbf{T}_i \in \mathit{SE}(3), \mathbf{K}_i \in \mathbb{R}^{3\times 3},i=1,2,\dots,N\}$. In this section, we will first introduce the overall \Acronym framework in \cref{sec:overall_framework}. \cref{sec:encoding} describes how we use geometric positional encoding to inject geometry information for image features and object queries implicitly. In \cref{sec:decoding} we propose making object queries view-conditioned so that 3D boxes are predicted in the specified view. Lastly, we present the novel viewpoint equivariance learning objective in \cref{sec:mtv}, which exploits the viewpoint-awareness of object queries and produces stronger geometric signals to improve 3D detection.

\subsection{Overall framework}
\label{sec:overall_framework}

The workflow of our proposed \Acronym builds upon a transformer-based architecture, as depicted in \cref{fig:overallarch}.
We first employ a backbone network that extracts image features $\{\mathbf{F}_i\in\mathbb{R}^{C\times H'\times W'}\}$ from multi-view images.
For each 2D location on the feature map grid, we calculate a geometric positional encoding that jointly considers pixel location, camera pose, and intrinsics. 

The image features, with their associated positional encoding, are flattened and processed by a transformer decoder \cite{carion2020end} with a set of object queries $\{\mathbf{q}_j\}$. The queries are constructed from a set of learnable 3D \textit{query points} $\{\mathbf{c}_j\}$ combined with a given \textit{query view} $\{\mathbf{T}^v\}$.

A series of self- and cross-attention layers then aggregate and update the 3D scene information into the queries, after which a feed-forward detection head maps the updated queries to box predictions $\{\mathbf{\hat{b}}_j\}$.
The box predictions are conditioned and expressed in the query views $\mathbf{T}^v_j$ associated with the queries, as detailed in \cref{sec:decoding}. 
Finally, we optimize the network by applying a viewpoint equivariance (VE) loss on the view-conditioned box predictions, as detailed in Sec~\ref{sec:mtv}.

\input{figures/fig2-overall_architecture.tex}

\subsection{Geometric positional encoding}
\label{sec:encoding}

Positional encodings provide location information of feature embeddings in Transformer architectures \cite{vaswani2017attention}. In this work, we encode the 3D geometric attributes associated with the image features as well as the object queries, when they are processed by the decoder. Inspired by~\cite{yifan2022input,liu2022petr}, for the image features we propose to encode the camera pose and the 3D inverse projection ray that combines pixel position and camera perspective geometry; for object queries the learnable 3D query point and the selected query view are encoded (more in \cref{sec:decoding}).

Specifically, given the extracted image features $\{\mathbf{F}_i\}$, we construct a triplet of geometric attributes, $[\mathbf{r}_{(u_i,v_i)}, \Quat_i,\mathbf{t}_i]$ for \textit{each feature location} $(u_i,v_i)$. 
$[\Quat_i,\textbf{t}_i]$ denote the quaternion vector and translation of the camera pose, and 
$\mathbf{r}_{(u_i,v_i)}$ denotes a unit-length inverse perspective projection ray originating from the pixel location given by:

\begin{equation}
\label{eq:rays}
    \mathbf{r}_{(u_i,v_i)}'=(\mathbf{K}_i\mathbf{R}_i^T)^{-1}[\alpha u_i,\alpha v_i,1]^T, \mathbf{r}=\frac{\mathbf{r}'}{||\mathbf{r}'||_2},
\end{equation}
where $\alpha$ is the downsample factor of $\mathbf{F}_i$ compared to image $\mathbf{I}_i$, $\mathbf{K}_i$ and $\mathbf{R}_i$ are instrinsic and rotation matrix of camera $i$.
The triplet $[\mathbf{r}_{(u_i,v_i)}, \Quat_i,\mathbf{t}_i]$ fully describes the perspective geometry for a given image feature $\mathbf{F}_i(u_i,v_i)$. Compared to PETR \cite{liu2022petr,liu2022petrv2} which model the positional information of image features by manually sampling a set of 3D point locations along the ray at pre-defined depth frustums, \Acronym employs a simpler design\footnote{PETR also combines a few other components with the 3D PE, namely 2D grid PE and view number PE, which we do not use.} and chooses not to assume the discretized depth prior, as we believe $[\mathbf{r}_{(u_i,v_i)}, \Quat_i,\mathbf{t}_i]$ keeps the full geometry information with which the model can learn 3D localization better.

\mypar{Learnable geometry mapping} 

We encode the geometric attributes into high-dimensional embeddings via Fourier transform followed by a learnable mapping.
Inspired by advances in NeRF \cite{mildenhall2021nerf,tancik2020fourier}, we first apply a Fourier transform to capture the fine-grained changes in the geometric attributes.
\begin{equation}
\label{eq:input_fourier}
   \gamma(x|[f_1,\dots,f_k]) = [\sin{(f_1\pi x)},\cos{(f_1\pi x)},\dots]
\end{equation}
The $k$ frequencies $[f_1,\dots,f_k]$ are sampled evenly between $[0, f_{\mathrm{max}}]$. Afterward, an MLP is used to project the output to dimension $C$ as our final geometric positional encoding:
\begin{equation}
\label{eq:input_embed}
    \mathbf{p}_{(u_i,v_i)}^e = \mathrm{MLP}_{\mathrm{enc}}(\gamma([\mathbf{r}_{(u_i,v_i)}, \Quat_i,\mathbf{t}_i])
\end{equation}

As a result, even without explicitly projecting the image features $\{\mathbf{F}_i\}$ back to 3D space, they become 3D geometry-aware when augmented with the 3D geometric positional encodings $\{\mathbf{P}_i^e \in \mathbb{R}^{C\times H'\times W'}\}$. Hence, we implicitly encode the multi-view perception of the scene at an input level, which will work jointly with our proposed VE learning objective to enforce 3D geometric modeling. 

\mypar{Temporal modeling} In the context of a multi-sweep setting, we follow \cite{liu2022petrv2} and transform the camera pose from previous frames into the current global frame via ego-motion compensation. The multi-sweep features are concatenated at the token dimension.

\subsection{View-conditioned query}
\label{sec:decoding}

\Acronym adopts a DETR-style \cite{carion2020end} decoder that consists of $L$ transformer layers, as shown in \cref{fig:overallarch}. Each layer performs self-attention among a set of $M$ queries $\{\mathbf{q}_j \in \mathbb{R}^C, j=1,2,\dots,M\}$, and cross-attention between the queries and the 3D geometry-aware image features $\{(\mathbf{F}_i, \mathbf{P}_i^e)\}$. The updated queries $\{\mathbf{q}_j\}$ will serve as input to the next layer:
\begin{equation}
\label{eq:decoding}
\resizebox{.9\hsize}{!}{$
    \{\mathbf{q}_j\}_{l} = \psi_{l-1}(\{\mathbf{F}_i\}, \{\mathbf{P}_i^e\}, \{\mathbf{q}_j\}_{l-1}), l=1,2,\dots,L$,}
\end{equation}
where $L$ is the number of attention layers. 
A classification and regression MLP heads map the queries from each layer into class logits and bounding box predictions, respectively.

\begin{equation}
\label{eq:output_heads}
    \mathbf{\hat{s}}_j = \mathrm{MLP}_{\mathrm{cls}}(\mathbf{q}_j), \mathbf{\hat{b}}_j = \mathrm{MLP}_{\mathrm{reg}}(\mathbf{q}_j)
\end{equation}

We propose to ground the queries with multi-view geometry. Concretely, a query $\mathbf{q}_j$ is constructed from two parts: a \textit{3D query point} and a \textit{query view}. We initialize a set of $M$ learnable 3D query points $\{\mathbf{c}_j\in\mathbb{R}^3, j=1,2,\dots,M\}$ in the \textit{global frame} $\mathbf{T}^0$, similarly to PETR~\cite{liu2022petr}, which is optimized during training.

\mypar{Query views} For the second part of query geometry, a query view $\mathbf{T}^v=[\Quat^v,\mathbf{t}^v]$ is selected relative to the global frame.
To construct the query, the 3D query points are first transformed into the query view via $\mathbf{c}_j^v=(\mathbf{T}^v)^{-1}\mathbf{c}_j$, and together with the query view $[\mathbf{c}_j^v,\Quat^v,\mathbf{t}^v]$ compose the query geometries. As described in \cref{sec:encoding}, the query geometries are similarly mapped by a Fourier transform followed by an MLP, into \textbf{view-conditioned queries}:

\begin{equation}
\label{eq:queries}
    \mathbf{q}_j^v=\mathrm{MLP}_{\mathrm{dec}}(\gamma([\mathbf{c}_j^v,\Quat^v,\mathbf{t}^v])).
\end{equation}

For query views, we refer the \textit{global frame} $\mathbf{T}^0=[[1,0,0,0],\mathbf{0}]$ as a default query view.\footnote{
This view is also the egocentric coordinate frame for box prediction and evaluation as done in other multi-view detection works~\cite{wang2022detr3d,liu2022petr,liu2022petrv2}.}
Additionally, we generate $V$ \textit{virtual query views} to provide variation to the decoding views and encourage viewpoint awareness in the model. Concretely, we randomly sample Euler angles $\Theta^v\in\mathbb{R}^3$ and translation $\mathbf{t}^v\in\mathbb{R}^3$ from uniform distributions $[\Theta_{\mathrm{min}}, \Theta_{\mathrm{max}}]$ and $[\mathbf{t}_{\mathrm{min}}, \mathbf{t}_{\mathrm{max}}]$, after which the Euler angles will be converted to the equivalent quaternion $\Quat^v\in SO(3)$, giving $\{\mathbf{T}^v=[\Quat^v,\mathbf{t}^v]\}$. In total, there are $V+1$ query views consisting of the default view and $V$ virtual views $\{\mathbf{T}^v, v=0,1,\dots,V\}$. Therefore, given the $M$ 3D query points $\{\mathbf{c}_j\}$ and $V+1$ query views $\{\mathbf{T}^v\}$, we construct object queries from $\{\mathbf{c}_j\}\times \{\mathbf{T}^v\}$, resulting in total $M\times(V+1)$ individual object queries.

\mypar{View-conditioned predictions} The query view specifies the coordinate system in which boxes (groundtruth, predicted) are defined. Specifically, given a \textit{view-conditioned} query $\mathbf{q}_j^v$, the box predictions $\mathbf{\hat{b}}_j^v$ are local to the underlying query view $\mathbf{T}^v$, parameterized as:
\begin{equation}
     \mathbf{\hat{b}}_j^v=[\Delta\mathbf{\hat{c}}_j^v, \mathbf{\hat{d}}_j, \cos(\phi), \sin(\phi), \mathbf{\hat{v}}_j^v] ,  
\end{equation}
where $\Delta\mathbf{\hat{c}}_j^v\in\mathbb{R}^3$ is the offset from the 3D query point $\mathbf{c}_j^v$ to the bounding box center, $\mathbf{\hat{d}}_j\in\mathbb{R}^3$ is the box dimensions, $\phi$ is the yaw angle of the box, and $\mathbf{\hat{v}}_j^v\in\mathbb{R}^3$ is the box velocity. 
As for the object classification score, we simply decode one from the global frame for each query point $\textbf{c}_j$, as it is simple and decoding from virtual views did not show advantage in our experiments. We predict a binary score for each class normalized by a sigmoid function.

The view-conditioned queries and their local predictions serve as a form of data augmentation during training and, more importantly, enable viewpoint equivariance regularization as discussed in \cref{sec:mtv}. More design choices are also ablated in \cref{sec:ablations}.
We only use the global frame $\mathbf{T}^0$ as the query view at inference time.

\subsection{Viewpoint equivariance loss}
\label{sec:mtv}

\input{figures/fig3-multivew.tex}

As described in \cref{sec:decoding}, given $V+1$ query views, there are $V+1$ versions of bounding box predictions $\{\mathbf{\hat{b}}_j^v\}$ coming from a single query point $\mathbf{c}_j$. 
The $V+1$ bounding boxes are expressed in different coordinate frames but of the same underlying ground truth object.
According to multi-view geometry, the observations of the \textit{same} object from different frames should be geometrically consistent and only differ by the relative transformation as shown in \cref{fig:mtvtarget}. 
 Therefore, we propose a viewpoint equivariance objective that considers the multi-view predictions coming from the same query point $c_j$ and box target from all query views \textit{jointly}. 
 
Concretely, we first ensure that the $V+1$ versions predictions from query point $\mathbf{c}_j$ are assigned to the same ground truth object. 
To achieve this goal, we create a \textbf{super box} by concatenating the predictions from different query views:
\begin{equation}
    \mathbf{\hat{B}}_j=[\mathbf{\hat{b}}_j^0,\mathbf{\hat{b}}_j^1,\dots,\mathbf{\hat{b}}_j^V].
\end{equation}
Similarly, we extend all the ground truth bounding boxes into super boxes:
\begin{equation}
    \mathbf{{B}}_m=[\mathbf{{g}}_m^0,\mathbf{{g}}_m^1,\dots,\mathbf{{g}}_m^V],
\end{equation}
where $\mathbf{g}^v_m$ is the ground truth bounding box in the expressed in the query view $\{\mathbf{T}^v\}$.
 
 Next, we perform Hungarian matching \cite{kuhn1955hungarian} to decide the optimal assignment between $\{\mathbf{\hat{B}}_j\}$ and $\{\mathbf{B}_m\}$, using the following cost function similar to DETR \cite{carion2020end}:

 \begin{equation}
\label{eq:loss}
\resizebox{.87\hsize}{!}{
  $  {\sigma} = -\mathds{1}_{\{c_m\neq \emptyset\}}\log (\textbf{s}_j(c_m)) + \mathds{1}_{\{c_m\neq \emptyset\}}L_{reg}(\mathbf{B}_m, \mathbf{\hat{B}}_j)$,}
\end{equation}
where $c_m$ is the ground truth class label and $L_{reg}()$ is a weighted L1 loss, given by:
\begin{align}
\label{eq:loss_reg}
\resizebox{.87\hsize}{!}{$
    L_{reg}(\mathbf{B}_m, \mathbf{\hat{B}}_j)=||\mathbf{\hat{b}}^0_j-\mathbf{g}^0_m||_1 + \Sigma_1^V\lambda_v ||\mathbf{\hat{b}}^v_j-\mathbf{g}^v_m||_1$.}
\end{align}
We use $\lambda_v$ to weigh the virtual views. Once we identify the optimal assignment, we calculate the loss on the super boxes:
\begin{equation}
\label{eq:loss_ve}
    L_{VE} = \lambda_{cls}L_{cls}(\textbf{s},c) + \lambda_{reg}L_{reg}(\mathbf{B}, \mathbf{\hat{B}}),
\end{equation}
for each paired prediction and ground truth. We adopt focal loss \cite{lin2017focal} for classification loss $L_{cls}$, and the same form of regression loss $L_{reg}$ as in matching. $\lambda_{cls}$ and $\lambda_{reg}$ are loss weights. 
For each 3D query point, by considering $V+1$ versions of predictions \textit{jointly} during both matching and optimization, the model learns viewpoint equivariance through multi-view consistency, leading to better 3D detection.

%% file: figures/fig2-overall_architecture.tex
\begin{figure*}[t!]
    \centering
    \includegraphics[width=0.9\textwidth]{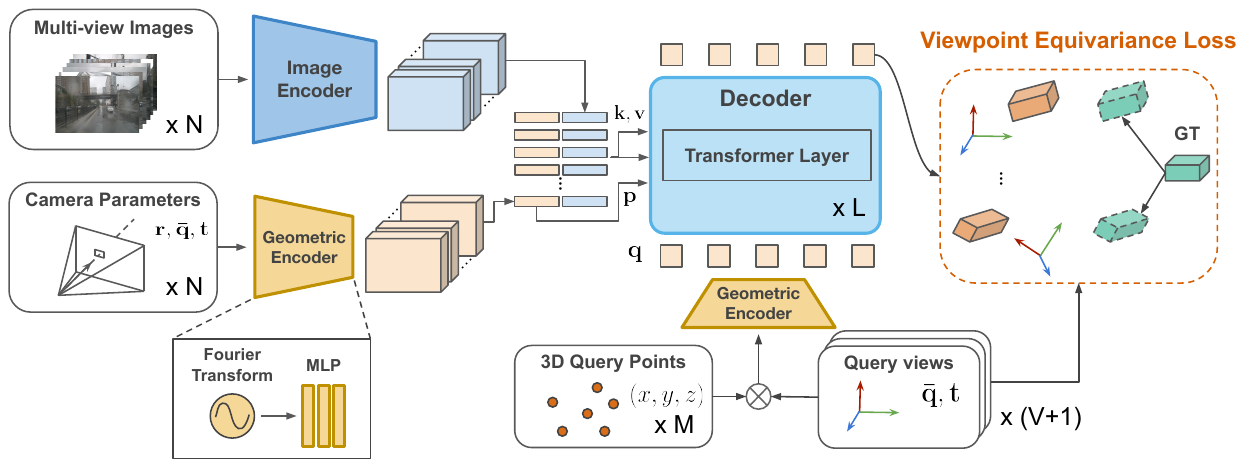}
    \caption{\textbf{The framework of our proposed \Acronym}: Given $N$ multi-view input cameras, an image encoder first extracts image features. For each feature embedding, we provide geometric positional encoding (PE) based on pixel location as well as camera geometrics (\cref{sec:encoding}). At the decoding stage, we apply a view-conditioned query constructed by 3D query points and Query Views to predict view-conditioned predictions (Sec.~\ref{sec:decoding}). Finally, we optimize the network through a novel viewpoint equivariance loss (Sec.~\ref{sec:mtv}).}
    \label{fig:overallarch}
\end{figure*} 

%% file: figures/fig3-multivew.tex
\begin{figure}
    \centering
    \includegraphics[width=0.85\linewidth]{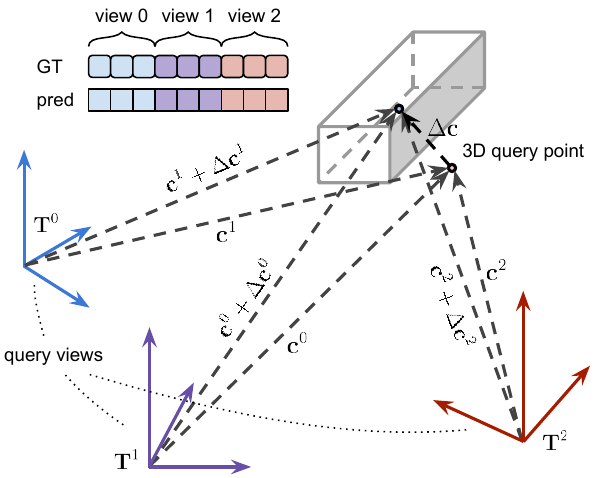}
    \caption{\textbf{Multi-view consistency enforced at training time.} According to multi-view geometry, the observations of the same box in different views should only differ by the relative transformation. Therefore, when a 3D query point is paired with multiple query views (3 in this illustration) to construct queries, the predictions in each respective view are \textit{combined} to match and regress the ground truth counterparts \textit{jointly}.
    }
    \label{fig:mtvtarget}
\end{figure}

%% file: sections/4_experiments.tex
\input{tables/tab1-nusc_val.tex}

\input{tables/tab2-nusc_test.tex}

\input{tables/tab3_ablations.tex}

\subsection{Experimental setup}
\mypar{Dataset and metrics} We evaluate our method on the large-scale benchmark NuScenes. NuScenes has 1000 scenes split into 700/150/150 as train/val/test subsets. Each scene contains 20-second videos collected by 6 surround-view cameras at 10Hz, with synchronized 3D box annotations at 2Hz. We report on metrics defined by NuScenes: mean Average Precision (mAP), mean Average Translation Error (mATE), mean Average Scale Error (mASE), mean Average Orientation Error, mean Average Velocity Error (mAVE), mean Average Attribution Error (mAAE), and NuScenes Detection Score (NDS) which is a comprehensive score that aggregates the above five sub-metrics.

\mypar{Implementation details} We adopt ResNet \cite{he2016deep} and VoVNetV2 \cite{lee2020centermask} with FPN \cite{lin2017feature} as the backbone network, and use the P4 features (1/16 image size) from the pyramid for all our experiments. We use the AdamW optimizer \cite{loshchilov2017decoupled} with cosine annealing \cite{loshchilov2016sgdr} to train \Acronym with learning rate starting at $2\times10^{-4}$ and ending at $2\times10^{-7}$, on 8 Tesla A100 GPUs with a total batch size 8. For image data augmentation, we apply random resize, horizontal flip, and crop; and for the 3D space we apply random scaling and rotation following CenterPoint \cite{yin2021center}. Importantly, when flipping the input image, we flip the 3D box annotations and camera extrinsics accordingly. The algorithm-specific hyper-parameters are ablated in \cref{sec:ablations} and fixed for all experiments. Please see the supplemental material's full list of hyper-parameters and more training details. All experiments are trained for 24 epochs except for test submission, which is trained for 96 epochs without CBGS \cite{zhu2019class}.

\subsection{Comparison to state of the art}

Our \Acronym achieves state-of-the-art detection performance on NuScenes val set across a range of model setups as shown in \cref{tab:1_nusc_val}, compared to previous works and some concurrent preprints~\cite{liu2022petrv2,jiang2022polarformer}. We use ImageNet-pretrained models for setups with ResNet-50/101 backbones to process image resolutions of $384\times1056$ and $512\times1408$, and outperform existing baselines. As the full-fledged setup, we adopt a V2-99 backbone initialized with depth-pretrained weights \cite{park2021pseudo}, operating on $640\times1600$ images. In this high-performance regime we compare two closely related works, PETR \cite{liu2022petr}, and PETRv2 \cite{liu2022petrv2}, as shown in the third group. \Acronym surpasses PETRv2 by 2.0$\%$ mAP and 1.0\% NDS, excelling at 3 sub-metrics. Our \textit{single-frame} version \Acronym-SF also achieves substantial gain over the single-frame baseline PETR, by 2.9\% mAP and 3.9\% NDS; it even outperforms the two-frame PETRv2 by 0.2\% mAP and at 3 sub-metrics. The noticeably lower mATE scores of \Acronym and \Acronym-SF further verify the strong localization capability.
For the test submission we adopt the depth-pretrained V2-99 backbone from ~\cite{park2021pseudo} with $640\times1600$ images. Without using the more advantageous data sampling strategy CBGS \cite{zhu2019class} as all the other baselines do, \Acronym still outperforms PETRv2 and achieves state-of-the-art performance with 50.5\% mAP and 58.5\% NDS.

\subsection{Ablation studies}
\label{sec:ablations}

 We first ablate the choices of some important hyper-parameters used in \Acronym shown in \cref{tab:ablate_freq,tab:ablate_bands,tab:ablate_views,tab:ablate_weights}, and then analyze the most critical components proposed in \Acronym by adding one component at a time as shown in \cref{tab:ablate_component_add} or making variations of specific components shown in \cref{tab:ablate_component_var}.

\mypar{Hyper-parameter selection.} We ablate the maximum Fourier frequency $f_{\mathrm{max}}$ \cref{tab:ablate_freq} and number of Fourier bands $k$ \cref{tab:ablate_bands} used in \cref{eq:input_fourier,eq:queries}, the number of virtual query views $V$ \cref{tab:ablate_views} and their weighting $\lambda_{\mathrm{v}}$ \cref{tab:ablate_weights} used in box regression loss. The table shows that \Acronym is robust across a wide range of hyper-parameter choices with competitive performance. Importantly, in \cref{tab:ablate_views} starting from not using any virtual query views ($V=0$), adding more views gradually improves performance until too many views might have caused the optimization to be more difficult. We choose $f_{\mathrm{max}}=8,k=64,V=2,\lambda_{\mathrm{v}}=0.2$ according to the best results and fix them for all experiments.

\mypar{Geometric positional encodings and object queries.} In \cref{tab:ablate_component_add}, we note that \#1 is effectively PETR \cite{liu2022petr}. The ``Fourier+MLP GPE'' column means we switch the position embeddings and queries in PETR with ours introduced in \cref{sec:encoding,sec:decoding}. From \#1 to \#2 we can see  significant improvements in mAP by 1.6\% and NDS by 1.7\%, which demonstrates that \textit{our proposed implicit geometric mapping better captures the 3D geometries} thanks to its Fourier component and the use of geometric attributes $[\mathbf{r}, \Quat,\mathbf{t}]$.

\mypar{Virtual query views.} In \cref{tab:ablate_component_add}, we then add $V=2$ virtual views during training on top of adopting our proposed geometric position embeddings and object queries, for both single-frame and 2-frame (``2-frame'' column) settings. From \#2 to \#3 , we see further jumps in mAP by 1.4\% and NDS by 2.4\%; from \#4 to \#5, mAP increases by 1.9\% and NDS by 3.2\%. This shows \textit{our proposed multi-view consistency loss applied on virtual view decoding effectively guides the model to improve 3D detection}. 

\mypar{Mutual benefit between \Acronym and multi-sweep.} Based on the above two comparisons, another observation is that our proposed \Acronym benefits in the 2-frame setting more than the single-frame setting, indicating that more geometric cues can be exploited when the input images contain richer multi-view context. Similarly, when looking at \#2 to \#4 (+1.2\% mAP, +3.1\% NDS) and \#3 to \#5 (+1.7\% mAP, +3.9\% NDS), we can see that adding more frames becomes more helpful after we add in the viewpoint equivariance on $V=2$ views. These two observations further consolidate the effectiveness of our geometric position embeddings, object queries, and virtual views.

\mypar{Fourier encoding.} In \cref{tab:ablate_component_var} ``no Fourier'' we show the importance of the Fourier encoding before the MLP by simply dropping it for both position embeddings and object queries, such that the MLPs map the geometric primitives directly to the 256-dim vectors. This leads to a drastic decrease in the detection performance (-6.9\% mAP and -5.1\% NDS), showing the critical role of the Fourier encoding to capture fine-grained changes in the geometries, which can be considered as high-frequency signals \cite{tancik2020fourier}.

\mypar{Partial camera geometry.} In \cref{tab:ablate_component_var} ``no $\mathbf{\bar{q}}$'' and ``no $\mathbf{t}$'' we show that removing partial information from the input cameras' poses leads to noticeably degraded performance. Specifically, removing the rotation information leads to drops in mAP by 1.5\% and NDS by 1.2\%, since the rotation indicates how the perspective projection plane is facing, which the rays are insufficient to describe; removing the translation leads to catastrophic drops in mAP by 9.2\% and NDS by 7.4\%, justifying the importance of translation information which is too difficult to infer from data implicitly if missing.

\mypar{Multi-view consistency.} In \cref{sec:mtv}, we constrain the different views of a query point to be considered simultaneously by concatenating the box predictions for matching and calculating loss. Instead, in \cref{tab:ablate_component_var} ``no joint match'' we treat them as individual objects. This leads to the possibility that a query point with different query views can be matched to different boxes, which leads to $2.6\%$ drop in mAP and 3.9\% drop in NDS. This demonstrates the multi-view consistency of the \textit{same} query point is meaningful, whereas simply augmenting the queries with views and treating all queries individually does not exploit the geometric signals enough. Without multi-view consistency, the excessive number of queries might be even harmful to the optimization, as indicated by the lower performance (42.5\% mAP, 48.3\% NDS) than the ``$V=0$'' VEDet (43.2\% mAP, 49.5\% NDS) in \cref{tab:ablate_views}.

\subsection{More analysis on viewpoint equivariance}

\input{figures/fig4-one-to-many.tex}

Given our design of view-conditioned queries and utilizing multiple virtual views during training, a natural question arises: does the performance gain come from the viewpoint equivariance regularization, or simply because more queries participate in the training? To show the effectiveness of the viewpoint equivariance, we compare to an intuitive baseline described as follows. We start from a plain version where we do not apply virtual views in queries (i.e. setting $V=0$). Under this setting, we add an additional set of $V\times M$ query points during training while duplicating the box targets by $V$ more times, and only use the original $M$ queries during inference. This setting \textit{matches the number of participating queries and box targets as \Acronym but does not contain any viewpoint equivariance regularization}, denoted by ``+2$M$ qry, no VE..'' in \cref{fig:one-to-many}, for NuScenes val set performances.

As shown in \cref{fig:one-to-many}, the extra queries and box targets generate more learning signals at the early stage as reflected by the superior mAP and NDS (\textcolor{orange}{orange curves}) compared to our ``$V=0$'' VEDet (\textcolor{blue}{blue curves}). However, their effects diminish as the curves plateau when reaching the end, and eventually ``+2$M$ qry, no VE.'' underperforms ``VEDet-0 views'', as shown in both plots of \cref{fig:one-to-many}. In contrast, Our counterpart VEDet with 2 virtual views during training, denoted by ``VEDet, $V=2$'' (\textcolor{purple}{purple curve}), outperforms ``VEDet, $V=0$'' consistently throughout the training and noticeably boosts the performance by +1.9\% mAP and +3.2\% NDS to 45.1\% mAP to 52.7\% NDS. This justifies the viewpoint equivariance regularization is more than just increasing the number of queries, and that it brings richer geometric learning signals for the model.

%% file: tables/tab1-nusc_val.tex
\begin{table*}[t!]
\centering
\resizebox{1.95\columnwidth}{!}{
\begin{tabular}{c|c|c|c|c|c|cccccc}
    \Xhline{1pt}
    Method & Backbone & Image size & CBGS & mAP$\uparrow$ & NDS$\uparrow$ & mATE$\downarrow$ & mASE$\downarrow$ & mAOE$\downarrow$ & mAVE$\downarrow$ & mAAE$\downarrow$ \\
    \hline 
    PETR & Res-50 & 384$\times$1056 & \checkmark & 0.313 & 0.381 & 0.768 & \textbf{0.278} & 0.564 & 0.923 & 0.225 \\
    \rowcolor{lightgray} \Acronym & Res-50 & 384$\times$1056 & \checkmark  & \textbf{0.347} & \textbf{0.443} & \textbf{0.726} & 0.282 & \textbf{0.542} & \textbf{0.555} & \textbf{0.198} \\
    \Xhline{1pt}
    FCOS3D & Res-101-DCN & 900$\times$1600 &  & 0.295 & 0.372 & 0.806 & 0.268 & 0.511 & 1.131 & \textbf{0.170} \\
    DETR3D$\dagger$ & Res-101-DCN & 900$\times$1600 & \checkmark & 0.349 & 0.434 & 0.716 & 0.268 & 0.379 & 0.842 & 0.200 \\
    BEVFormer$\dagger$ & Res-101-DCN & 900$\times$1600 &    & 0.416 & 0.517 & 0.673 & 0.274 & 0.372 & \textbf{0.394} & 0.198 \\
    UVTR$\dagger$ & Res-101-DCN & 900$\times$1600 &    &  0.379 & 0.483 & 0.731 & \textbf{0.267} & \textbf{0.350} & 0.510 & 0.200 \\
    PETR & Res-101 & 512$\times$1408 & \checkmark & 0.357 & 0.421 & 0.710 & 0.270 & 0.490 & 0.885 & 0.224 \\
    \rowcolor{lightgray} \Acronym & Res-101 & 512$\times$1408 &   \checkmark & \textbf{0.432} & \textbf{0.520} & \textbf{0.638} & 0.275 & 0.362 & 0.498 & 0.191\\ 
    \Xhline{1pt}
    PETR$\ddagger$ & V2-99 & 640$\times$1600 &   & 0.404  & 0.447 & 0.739 & 0.271 & 0.452 & 0.876 & 0.208 \\ 
    PETRv2$\ddagger$ & V2-99 & 640$\times$1600 &   & 0.431  & 0.517 & 0.730 & 0.264 & 0.399 & \textbf{0.404} & \textbf{0.190} \\ 
    \rowcolor{lightgray} \Acronym-SF$\ddagger$ & V2-99 & 640$\times$1600 &   & 0.433 & 0.486  & 0.683 & \textbf{0.263} & 0.352 & 0.808 & 0.201 \\ 
    \rowcolor{lightgray} \Acronym$\ddagger$ & V2-99 & 640$\times$1600 &   & \textbf{0.451}  & \textbf{0.527} & \textbf{0.670} & \textbf{0.263} & \textbf{0.347} & 0.510 & 0.192 \\

    \Xhline{1pt}
\end{tabular}
}
\caption{\textbf{NuScenes val set performance.} Our \Acronym outperforms existing baselines consistently cross various backbone and resolution choices. $\dagger$ means initializing from FCOS3D backbone. $\ddagger$ means initializing from the depth-pretrained backbone provided by DD3D \cite{park2021pseudo}.}
\label{tab:1_nusc_val}
\end{table*}

%% file: tables/tab2-nusc_test.tex
\begin{table*}[t!]
\centering
\resizebox{1.98\columnwidth}{!}{
\begin{tabular}{c|c|c|c|c|c|cccccc}
    \Xhline{1pt}
    Method & Backbone & Image size & TTA & mAP$\uparrow$ & NDS$\uparrow$ & mATE$\downarrow$ & mASE$\downarrow$ & mAOE$\downarrow$ & mAVE$\downarrow$ & mAAE$\downarrow$ \\
    \hline 
    DD3D$\ddagger$ & V2-99 & 900$\times$1600 & $\checkmark$ & 0.418 & 0.477 & 0.572 & 0.249 & 0.368 & 1.014 & 0.124 \\
    DETR3D$\ddagger$ & V2-99 & 900$\times$1600 & $\checkmark$ & 0.412 & 0.479 & 0.641 & 0.255 & 0.394 & 0.845 & 0.133  \\
    PETR$\ddagger$ & V2-99 & 640$\times$1600 &  & 0.434 & 0.481 & 0.641 & 0.248 & 0.437 & 0.894 & 0.143 \\
    UVTR$\ddagger$ & V2-99 & 900$\times$1600 &  & 0.472 & 0.551 & 0.577 & 0.253 & 0.391 & 0.508 & 0.123 \\
    BEVFormer$\ddagger$ & V2-99 & 900$\times$1600 &  & 0.481 & 0.569 & 0.582 & 0.256 & 0.375 & 0.378 & 0.126 \\
    BEVDet4D & Swin-B & 900$\times$1600 & $\checkmark$ & 0.451 & 0.569 & \textbf{0.511} & 0.241 & 0.386 & \textbf{0.301} & 0.121\\
    PolarFormer$\ddagger$ & V2-99 & 900$\times$1600 &  & 0.493 & 0.572 & 0.556 & 0.256 & 0.364 & 0.439 & 0.127 \\
    PETRv2$\ddagger$ & V2-99 & 640$\times$1600 &  & 0.490 & 0.582 & 0.561 & \textbf{0.243} & 0.361 & 0.343 & \textbf{0.120} \\
    \rowcolor{lightgray} \Acronym$\ddagger$ & V2-99 & 640$\times$1600 &  & \textbf{0.505} & \textbf{0.585} & 0.545 & 0.244 & \textbf{0.346} & 0.421 & 0.123 \\
    \Xhline{1pt}
\end{tabular}
}
\caption{\textbf{NuScenes test set performance.} Our \Acronym achieves state-of-the-art performance compared to existing publications. We also note that other baselines are trained with CBGS which is a more advantegeous sampling strategy. $\ddagger$ means initializing from the depth-pretrained backbone provided by DD3D \cite{park2021pseudo}.}
\label{tab:2_nusc_test}
\end{table*}

%% file: tables/tab3_ablations.tex
\begin{table*}
  \setlength\tabcolsep{4pt}
  \label{tab:ablations}
  \hfill%
  \subfloat[\textbf{Fourier frequency.}]{
    \label{tab:ablate_freq}
    \begin{tabular}{lcc}
    \toprule
     $f_{\mathrm{max}}$ &   mAP$\uparrow$ & NDS$\uparrow$ \\
    \midrule
       4 & 0.446 & 0.521 \\
       \rowcolor{lightgray} 8 & \textbf{0.451} & \textbf{0.527} \\
       16 & 0.447 & 0.521 \\
       32 & 0.442 & 0.517 \\
    \bottomrule
    \end{tabular}}
  \hfill%
  \subfloat[\textbf{\# Fourier bands.}]{
    \label{tab:ablate_bands}
    \begin{tabular}{lcc}
    \toprule
     $k$ &   mAP$\uparrow$ & NDS$\uparrow$ \\
    \midrule
       32 & 0.446 & 0.522 \\
       \rowcolor{lightgray} 64 & \textbf{0.451} & \textbf{0.527} \\
       96 & 0.445 &  0.520 \\
       128 & 0.438 &  0.512 \\
    \bottomrule
    \end{tabular}}
  \hfill%
  \subfloat[\textbf{\# virtual views}]{%
    \label{tab:ablate_views}
    \resizebox{0.31\columnwidth}{!}{
    \begin{tabular}{lcc}
    \toprule
     $V$ &   mAP$\uparrow$ & NDS$\uparrow$ \\
    \midrule
       0 & 0.432 & 0.495 \\
       1 & 0.441 & 0.518 \\
       \rowcolor{lightgray} 2 & \textbf{0.451} & \textbf{0.527} \\
       4 & 0.448 & 0.526 \\
       6 & 0.437 & 0.511 \\
    \bottomrule
    \end{tabular}}}
  \hfill%
  \subfloat[\textbf{virtual view weights.}]{
    \label{tab:ablate_weights}
    \begin{tabular}{lcc}
    \toprule
     $\lambda_{\mathrm{v}}$ &   mAP$\uparrow$ & NDS$\uparrow$ \\
    \midrule
       0.1 & 0.442 & 0.522 \\
       \rowcolor{lightgray} 0.2 & \textbf{0.451} & \textbf{0.527} \\
       0.4 & 0.439 & 0.517 \\
       0.6 & 0.431 & 0.499 \\
    \bottomrule
    \end{tabular}}
  \hfill%
  \vspace{12pt}
  \hfill
  \subfloat[\textbf{components addition.}]{
    \label{tab:ablate_component_add}
    \hspace{30pt}
    \resizebox{0.99\columnwidth}{!}{
    \begin{tabular}{lccccc}
    \toprule
     \#  & Fourier+MLP GPE. & 2 virtual views & 2-frame &  mAP$\uparrow$ & NDS$\uparrow$ \\
    \midrule
      1 & & & & 0.404 & 0.447 \\
      2 & \checkmark & & & 0.420 & 0.464 \\
      3 & \checkmark & \checkmark & & 0.434 & 0.488 \\
      4 & \checkmark & & \checkmark & 0.432 & 0.495 \\
     \rowcolor{lightgray} 5 & \checkmark & \checkmark & \checkmark & \textbf{0.451} & \textbf{0.527} \\
    \bottomrule
    \end{tabular}}}
  \hfill%
  \subfloat[\textbf{components variation}]{%
    \label{tab:ablate_component_var}
    \resizebox{0.49\columnwidth}{!}{
    \begin{tabular}{lcc}
    \toprule
     method &   mAP$\uparrow$ & NDS$\uparrow$ \\
    \midrule
       \rowcolor{lightgray} \Acronym & \textbf{0.451} & \textbf{0.527} \\
       no Fourier & 0.382 & 0.476 \\
       no $\mathbf{\bar{q}}$ & 0.436 & 0.515 \\
       no $\mathbf{t}$ & 0.353 & 0.453 \\
       no joint match & 0.425 & 0.483 \\
    \bottomrule
    \end{tabular}}}
  \hfill%
  \hfill
  \caption{\textbf{Ablation studies and analyses.} In \cref{tab:ablate_freq,tab:ablate_bands,tab:ablate_views,tab:ablate_weights} we first analyze important hyper-parameters relevant to the algorithm, and choose the ones giving the best performance. In \cref{tab:ablate_component_add} we ablate and show the importance of the proposed geometric embeddings and virtual views. They not only monotonically bring improvements to the 3D detection performance, but also mutually benefit adding time frames. In \cref{tab:ablate_component_var} we ablate some variations in the components, showing the critical usage of Fourier encoding and some optimal settings for queries during training and inference.}
  \end{table*}

%% file: figures/fig4-one-to-many.tex
\begin{figure}
    \centering
    \includegraphics[width=1.0\linewidth]{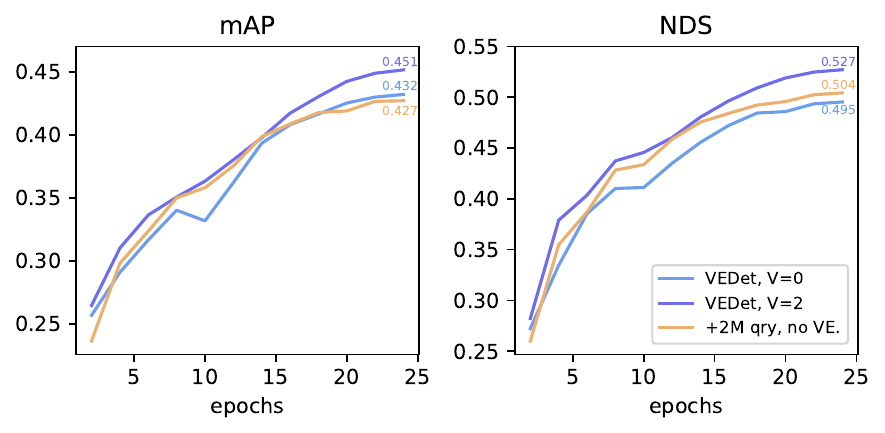}
    \caption{\textbf{Effectiveness of the view equivariant regularization.} Simply adding more queries and duplicating box targets without introducing view equivariance only helps model optimization at an early stage and does not help the final detection performance as shown by the \textcolor{orange}{orange curves} compared to the \textcolor{blue}{blue curves}. VEDet is able to leverage richer geometric signals brought by the view equivariance objective with the help of view-conditioned queries, which leads to a performance boost shown by \textcolor{violet}{purple curves}.
    }
    \vspace{-4mm}
    \label{fig:one-to-many}
\end{figure}

%% file: sections/5_limitations.tex
\vspace{-2mm}
\mypar{Camera parameter robustness.} \Acronym leverage implicit geometric encodings to learn 3D geometry in a data-driven way, therefore the robustness of \Acronym against camera parameters is critical and worth investigating in future works.
\mypar{Depth information.} This work mainly leverages geometric signals generated from the ground truth 3D bounding boxes and generic pose information of both cameras and object queries, while some concurrent works \cite{li2022bevdepth, huang2022bevdet4d} explicitly use depth to guide the 3D feature construction and hence the interaction between the heads and features. Incorporating depth information into the framework, such as enhancing the geometric positional encodings or guiding spatial attention, will be explored for future works.
\mypar{Temporal modeling.} The current VEDet follows existing works to simply concatenate features from multi-sweeps along the token dimension. While this is effective, the concatenation has two main issues: it throws away the temporal ordering, and has difficulty in modeling long sequences due to memory and computation constraints. Investigation on a better temporal modeling such as recurrent processing will be valuable for future works.

%% file: sections/6_conclusion.tex
In this work, we introduce a novel camera-based multi-view 3D object detection framework that learns from viewpoint equivariance regularization. 
\Acronym employs a transformer decoder with a set of view-conditioned queries to decode bounding boxes from image features with geometric positional encodings.
The view-conditioning of queries enables us to enforce viewpoint equivariance on predictions made from different viewpoints, and therefore generate richer geometric learning signals to guide the model in better understanding the 3D structure of the scene. \Acronym achieves state-of-the-art 3D detection performance, and we conduct extensive experiments to show the effectiveness of its components. We also point out a few meaningful limitations for future works.